\documentclass[runningheads]{llncs}
\usepackage[T1]{fontenc}
\usepackage{graphicx}
\usepackage{amssymb}
\usepackage{amsmath}

\usepackage{booktabs}
\usepackage{multirow}
\usepackage{rotating} 
\let\oldthebibliography\thebibliography
\renewcommand{\thebibliography}[1]{%
  \oldthebibliography{#1}%
  \setlength{\itemsep}{0pt plus 0.3pt}%
  \setlength{\parskip}{0pt}%
}
\begin{document}
\title{BandVQ: Band-Wise Vector-Quantized EEG Foundation Model}
%
%
\author{Jamiyan Sukhbaatar \and
Satoshi Imamura \and
Toshihisa Tanaka}
\authorrunning{J. Sukhbaatar et al.}
%
\institute{Tokyo University of Agriculture and Technology, Tokyo, Japan \and
National University of Mongolia, Ulaanbaatar, Mongolia\\
\email{tanakat@cc.tuat.ac.jp}\\}
\maketitle              
\begin{abstract}
A central challenge in electroencephalography (EEG) foundation modeling is learning transferable representations across recordings with diverse tasks, montages, references, and spectral characteristics. Existing masked modeling approaches often rely on broadband continuous patches or a single discrete representation, which may underrepresent frequency-specific activity. This paper proposes \textbf{BandVQ}, a band-wise vector-quantized EEG foundation model that decomposes EEG into delta, theta, alpha, beta, and gamma bands, trains an independent VQ-VAE tokenizer for each band, and pretrains a shared Transformer encoder on the resulting discrete VQ code indices. The encoder uses masked code tokens, quantized absolute log-power tokens, channel and temporal embeddings, and metadata prefix tokens representing reference, band, task family, and phase. Region-based masking is also introduced to reduce the trivial reconstruction of spatially adjacent electrodes. The model is pretrained on 71 public EEG corpora comprising over 9,200 subjects and 357,000 single-channel hours and evaluated on six subject-independent classification datasets. Under the current evaluation setting, the proposed model achieves strong transfer performance, with the highest reported results on three cognitive tasks and competitive performance on three motor imagery tasks.
\end{abstract}

\section{Introduction}

Electroencephalography (EEG) provides a noninvasive measure of neural activity and has become an important modality for brain-computer interfaces, neurological monitoring, cognitive assessment, and affective computing~\cite{lotte2018review}. Learning general-purpose EEG representations remains challenging because EEG recordings vary substantially across datasets, subjects, tasks, montages, references, sampling rates, and acquisition systems~\cite{melnik2017systems,saha2020intra}. Such heterogeneity weakens the transferability of task-specific models, which are typically optimized for a particular dataset or experimental condition~\cite{xu2020cross}.

Recent EEG foundation models commonly adopt self-supervised masked modeling, in which multi-channel EEG is divided into temporal or spatio-temporal patches, a subset of the input is masked, and a neural network is trained to reconstruct the missing content or predict its latent representation~\cite{ref_CBraMod,ref_LaBraM}. Two representative strategies have emerged. Continuous reconstruction methods preserve waveform-level detail, but sample-wise reconstruction losses, such as mean squared error, can overemphasize high-amplitude components and may bias learning toward dominant low-frequency activity, given the characteristic decrease in EEG spectral power with frequency. Discrete token methods, including vector-quantized modeling, provide stable classification-like objectives and align naturally with masked token pretraining, but a single discrete bottleneck can discard fine-grained waveform and spectro-temporal details. These limitations suggest that EEG masked modeling should preserve frequency-specific waveform structure while also retaining amplitude information and acquisition context.

EEG also presents modality-specific challenges that are particularly important for masked token pretraining. First, EEG power spectra exhibit strong frequency-dependent amplitude differences, with lower frequency activity typically dominating the signal energy~\cite{ref_power1}. This imbalance can lead to broadband tokenization underrepresenting high-frequency components unless the spectral structure is explicitly modeled. Second, neighboring EEG electrodes often capture highly correlated activity because of volume conduction and spatial smoothing~\cite{burle2015spatial}. A masked token model may therefore partially solve the pretext task by exploiting nearby electrodes rather than relying on broader spatial and temporal structures. Third, large-scale EEG corpora combine recordings from various experimental conditions. Information such as reference scheme, task family, and recording phase can help the model interpret the same signal pattern under different acquisition and experimental contexts, but such metadata must be incorporated without making the model overly dependent on dataset-specific identifiers~\cite{pernet2019eegbids}.

To address these challenges, we propose \textbf{BandVQ}, a band-wise vector-quantized, metadata-conditioned EEG foundation model. The proposed framework decomposes EEG into five canonical frequency bands---delta, theta, alpha, beta, and gamma---and trains a separate vector-quantized variational autoencoder (VQ-VAE)~\cite{van2017neural} for each band. The trained tokenizers convert band-limited waveforms into discrete code sequences. A shared Transformer encoder is then pretrained on these band-wise codes using masked code prediction. The model receives discrete code tokens, quantized band power tokens, channel and temporal position embeddings, and metadata prefix tokens that encode the recording context. To reduce trivial reconstruction from spatially adjacent channels, we further introduce region-based masking, which masks anatomically related channel groups rather than independent random electrodes.

The band-wise design preserves the advantages of discrete masked token pretraining while reducing the information loss caused by broadband quantization. Per-token RMS normalization stabilizes VQ-VAE training across bands, and quantized absolute log-power tokens preserve amplitude information that would otherwise be removed by normalization. The shared Transformer learns a unified representation across all frequency bands while retaining band-specific prediction heads. This design allows the model to exploit the common temporal and spatial structure in EEG recordings while preserving frequency-specific code vocabularies.

We pretrain the BandVQ model using the public EEG corpus comprising 71 datasets introduced in SingLEM~\cite{ref_SingLEM}, totaling more than 9,200 subjects and 357,000 single-channel hours. We evaluate the model on six subject-independent downstream classification datasets, including three motor imagery and three cognitive tasks. BandVQ is compared with published results reported in the SingLEM benchmark for SingLEM~\cite{ref_SingLEM}, BENDR~\cite{ref_BENDR}, BIOT~\cite{ref_BIOT}, LaBraM~\cite{ref_LaBraM}, and CBraMod~\cite{ref_CBraMod}. Although the comparison is not fully controlled because the proposed model is fine-tuned while the reported baselines use fixed-feature SVM classifiers, the results show strong downstream performance, and ablations indicate that metadata conditioning and band-power tokens improve generalization.


The main contributions of this study are summarized as follows:
\begin{itemize}
    \item We propose \textbf{BandVQ}, a band-wise vector-quantized EEG foundation model that couples five frequency-specific VQ-VAE tokenizers with a shared Transformer encoder under a masked-token pretraining objective.
    \item We introduce three complementary conditioning mechanisms tailored to EEG heterogeneity: metadata prefix tokens (reference, band, task family, phase), quantized absolute log-power tokens that restore amplitude information after RMS normalization, and region-based masking that suppresses reconstruction shortcuts from neighboring electrodes.
    \item We pretrain the model on 71 public EEG corpora comprising over 9,200 subjects and 357,000 single-channel hours, and evaluate it on six subject-independent downstream classification tasks. The results show strong downstream performance, including the highest accuracy on three cognitive tasks under the comparison setting considered in this study.
\end{itemize}

\section{Proposed Method}

\subsection{Overview}

\begin{sidewaysfigure}
    \centering
    \includegraphics[width=\textwidth]{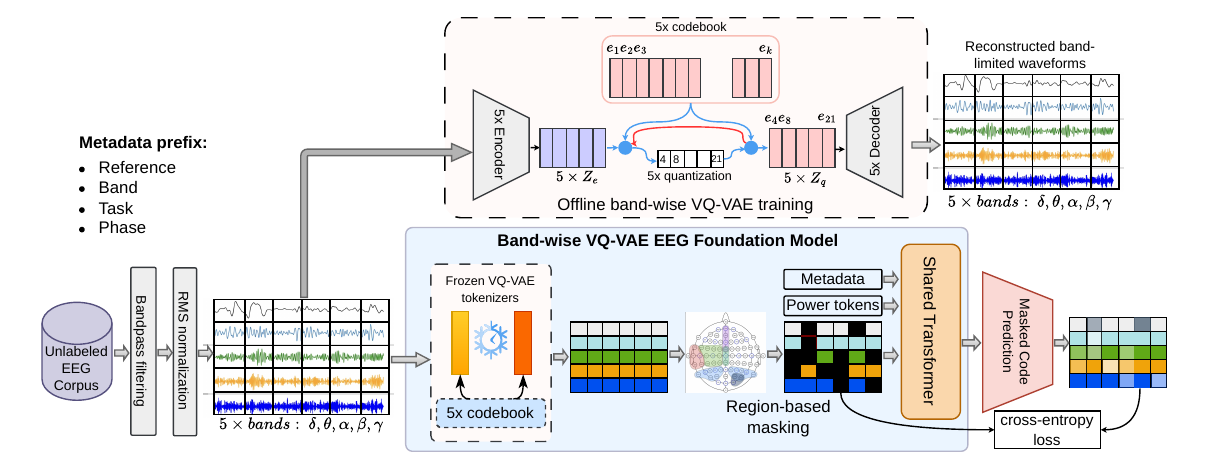}
    \caption{Overview of \textbf{BandVQ}, the proposed band-wise vector-quantized EEG foundation model. Band-specific VQ-VAE tokenizers are trained offline and then frozen to produce discrete code indices. A shared Transformer encoder is pretrained with masked code prediction using quantized absolute log-power tokens, metadata prefix tokens, and region-based masking.}
    \label{fig:architecture}
\end{sidewaysfigure}

The proposed framework learns transferable EEG representations through a two-stage self-supervised pipeline. First, EEG signals are decomposed into canonical frequency bands, and an independent vector-quantized variational autoencoder (VQ-VAE)~\cite{van2017neural} tokenizer is trained for each band. Each tokenizer converts normalized band-limited waveform tokens into discrete VQ code indices. Second, a shared Transformer encoder is pretrained on the resulting band-wise sequences of code indices using masked code prediction. The Transformer receives embeddings of masked code indices, quantized absolute log-power tokens, channel and temporal embeddings, and metadata prefix tokens that encode recording context. Fig.~\ref{fig:architecture} summarizes the two-stage framework, consisting of offline band-wise VQ-VAE tokenizer training and shared Transformer pretraining on masked discrete EEG code indices.


Let $\mathbf{X}\in\mathbb{R}^{C\times N}$ denote a multi-channel EEG segment with $C$ channels and $N$ time samples. During tokenizer training, each VQ-VAE processes a single-channel band-limited waveform token. During foundation model pretraining, the model processes multi-channel discrete token sequences from one sampled frequency band at a time. During downstream adaptation, all five band-specific token streams are extracted from each trial and encoded by the pretrained Transformer to obtain trial-level representations. Band-specific tokenizers preserve the spectral structure within each frequency range, whereas the shared Transformer learns common temporal, spatial, and contextual regularities across heterogeneous EEG recordings.

\subsection{Preprocessing and Band Decomposition}

All EEG recordings are processed with a common preprocessing pipeline prior to tokenization. The recordings are notch-filtered at the power-line frequency, bandpass-filtered to 0.5--45~Hz, resampled to a common sampling rate, and converted to microvolt units when necessary. Let $\mathbf{X}^{(\mu\mathrm{V})}$ denote the preprocessed EEG segment expressed in microvolts. To improve numerical conditioning during neural-network training, the signal is scaled by a constant factor $s$:
\begin{equation}
    \mathbf{X}_{\mathrm{scaled}}=\frac{\mathbf{X}^{(\mu\mathrm{V})}}{s}.
\end{equation}

Each preprocessed EEG channel is decomposed into five canonical frequency bands:
$\delta$ (0.5--4~Hz), $\theta$ (4--8~Hz),
$\alpha$ (8--13~Hz), $\beta$ (13--30~Hz),
and $\gamma$ (30--45~Hz).
Band decomposition is performed in the frequency domain for each channel. Before filtering, the signal is reflect-padded to reduce boundary artifacts. We then compute the real-valued fast Fourier transform of the padded signal, apply a band-specific frequency mask with cosine-tapered boundaries, and transform the filtered spectrum back to the time domain. The padded samples are cropped after filtering.

\subsection{Band-wise VQ-VAE Tokenization}

\subsubsection{Token Construction and RMS Normalization}

Each band-limited waveform is partitioned into non-overlapping temporal tokens. Let $L$ denote the length of the token in the samples. For the $n$-th token $\mathbf{x}_{n}\in\mathbb{R}^{L}$, the normalized token is given by
\begin{equation}
    \tilde{\mathbf{x}}_n=\frac{\mathbf{x}_n}{r_n}.
\end{equation}
where $r_n$ denotes the root mean square (RMS) value defined as
\begin{equation}
    r_n=\max\left(
    \sqrt{\frac{1}{L}\sum_{t=1}^{L}x_{n,t}^{2}+\epsilon},
    \rho
    \right),
\end{equation}
where $\epsilon$ is a stabilizing offset and $\rho$ is a minimum RMS floor.
RMS normalization stabilizes tokenizer training across frequency bands, where lower-frequency components typically dominate in amplitude. Because this normalization discards absolute amplitude information, we preserve it separately as a quantized log-power token provided to the Transformer, as described in the following subsection.

\subsubsection{Band-Specific VQ-VAE Tokenizers}

We train one VQ-VAE tokenizer for each band $b\in\{\delta,\theta,\alpha,\beta,\gamma\}$. Each tokenizer consists of a convolutional encoder, an exponential moving average vector quantizer, and a convolutional decoder. The encoder maps the normalized waveform token $\tilde{\mathbf{x}}_n$ to a latent vector $\mathbf{z}_{e,n}\in\mathbb{R}^{D}$. The quantizer assigns this vector to the nearest codebook entry:
\begin{equation}
    k_n=\arg\min_{k\in\{0,\dots,K_b-1\}}
    \left\|\mathbf{z}_{e,n}-\mathbf{e}^{(b)}_{k}\right\|_{2}^{2},
\end{equation}
where $K_b$ is the size of the codebook for band $b$ and $\mathbf{e}^{(b)}_{k}$ is the code vector with index $k$ in the corresponding band-specific codebook. The quantized vector is
\begin{equation}
    \mathbf{z}_{q,n}=\mathbf{e}^{(b)}_{k_n}.
\end{equation}
We use a straight-through estimator for the non-differentiable nearest-neighbor quantization step and update the codebook entries using an exponential moving average.

The decoder reconstructs the normalized waveform token from the quantized vector. Reconstructed tokens are concatenated in temporal order to form the normalized band-limited waveform reconstruction. The encoder and decoder are implemented with convolutional blocks, with downsampling in the encoder and upsampling in the decoder.

\subsubsection{Tokenizer Training Objective}

Each tokenizer reconstructs the normalized band-domain waveform. Let $\tilde{\mathbf{x}}$ denote the normalized target waveform, and $\hat{\mathbf{x}}$ denote the reconstruction. The tokenizer loss combines time-domain reconstruction, spectral reconstruction, and commitment terms:
\begin{equation}
    \mathcal{L}_{\mathrm{VQ}}
    =
    \lambda_{\mathrm{MAE}}\mathcal{L}_{\mathrm{MAE}}
    +
    \lambda_{\mathrm{MSE}}\mathcal{L}_{\mathrm{MSE}}
    +
    \lambda_{\mathrm{STFT}}\mathcal{L}_{\mathrm{STFT}}
    +
    \lambda_{\mathrm{commit}}\mathcal{L}_{\mathrm{commit}},
\end{equation}
where $\mathcal{L}_{\mathrm{MAE}}$ and $\mathcal{L}_{\mathrm{MSE}}$ are time-domain reconstruction losses, $\mathcal{L}_{\mathrm{STFT}}$ is a multi-scale STFT magnitude loss, and $\mathcal{L}_{\mathrm{commit}}$ is the VQ-VAE commitment loss.

\subsection{Power and Metadata Token Representation}

\subsubsection{Quantized Absolute Log-Power Tokens}

RMS normalization removes the absolute amplitude information from the waveform tokens. To preserve this information, we compute a band-power descriptor for each original unnormalized token. For token $\mathbf{x}_n$, the mean squared amplitude and logarithmic power are
\begin{equation}
    p_n = \frac{1}{L}\sum_{t=1}^{L}x_{n,t}^{2},
    \qquad
    a_n = \log_{10}\left(\max(p_n,\epsilon_p)\right),
\end{equation}
where $\epsilon_p$ prevents numerical instability near zero. The scalar $a_n$ is clipped to a fixed range $[a_{\min}, a_{\max}]$ and uniformly quantized into $B_p$ bins. The same clipping range and quantization scheme are used during pretraining and downstream preparation. This consistency gives the power embedding the same interpretation during self-supervised pretraining and supervised adaptation.

\subsubsection{Metadata Prefix Tokens}

Large-scale EEG corpora combine recordings acquired under different references, task paradigms, and experimental phases. To represent this context, each pretraining sample is associated with reference identifier $r$, frequency band identifier $b$, task family identifier $q$, and phase identifier $\phi$. If a metadata field is unavailable, a dedicated unknown identifier is used.

The task family identifier follows a fixed global taxonomy of coarse categories, including resting or baseline recordings, cognitive tasks, language tasks, sensorimotor activity, BCI paradigms, sensory perception, affective or social tasks, real-world mobility, clinical EEG, artifact labels, sleep, unknown events, and reserved categories. Dataset-specific event names are assigned to this taxonomy using a prioritized rule-based procedure.

Each sequence is also assigned a phase label that describes its temporal role within a trial or recording segment, such as baseline, cue, task, feedback, rest, or unknown. When fine-grained phase annotations are unavailable, the closest available phase is assigned or the unknown label is used.

\subsection{Region-Based Masking}

Standard independent token masking may allow the model to infer a masked EEG token from spatially neighboring electrodes. To reduce this shortcut, we use region-based masking based on channel-region annotations. The channels are grouped by anatomical region and laterality. Anatomical regions include frontal, central, temporal, parietal, and occipital areas. Laterality is defined by the standard electrode-name convention: odd-numbered electrodes correspond to the left hemisphere, even-numbered electrodes correspond to the right hemisphere, and electrodes with suffix ``z'' to the midline.

At each time index, the masking module samples region groups and masks the corresponding EEG tokens from channels whose annotations overlap with the sampled groups. Let $\mathcal{G}_{\mathrm{lat}}$ and $\mathcal{G}_{\mathrm{mid}}$ denote the available lateral and midline region groups in a sampled window, respectively. Given a masking ratio $p_{\mathrm{mask}}$, the numbers of sampled groups are
\begin{equation}
    k_{\mathrm{lat}}=\left\lceil p_{\mathrm{mask}}|\mathcal{G}_{\mathrm{lat}}|\right\rceil,
    \qquad
    k_{\mathrm{mid}}=\left\lceil p_{\mathrm{mask}}|\mathcal{G}_{\mathrm{mid}}|\right\rceil.
\end{equation}

The masking operation replaces the discrete code index with a dedicated mask-code identifier and replaces the corresponding power token with a dedicated mask-power identifier. The pretraining loss is computed only at the channel-time token locations selected for masking.

\subsection{Shared Transformer Encoder}

The shared Transformer encoder operates on one frequency band at a time. For an EEG segment with $C$ channels and $N$ time samples, let $T=N/L$ denote the number of non-overlapping temporal tokens per channel, where $L$ is the token length. Here, segments are cropped so that $N$ is divisible by $L$. For a sampled frequency band $b\in\{\delta,\theta,\alpha,\beta,\gamma\}$, the corresponding band-wise code indices are flattened in time-major order, giving an EEG-token sequence length $S=CT$. The flattened index $i\in\{1,\dots,S\}$ corresponds to one channel-time token, with channel identifier $\chi_i$ and temporal-position index $\tau_i$.

For each position $i$ of the EEG-token in this band-specific sequence, the input embedding is formed by summing the embedding of the masked code index, the embedding of the masked power token, the channel embedding, and the temporal-position embedding:
\begin{equation}
    \mathbf{h}^{(0)}_i
    =
    \mathbf{e}^{\mathrm{code}}(c_i^{m})
    +
    \mathbf{e}^{\mathrm{power}}(u_i^{m})
    +
    \mathbf{e}^{\mathrm{ch}}(\chi_i)
    +
    \mathbf{e}^{\mathrm{time}}(\tau_i),
\end{equation}
where $c_i^{m}$ and $u_i^{m}$ denote the code index and the power token after masking, respectively.

Four metadata prefix tokens representing the reference, frequency band, task family, and phase are prepended to the sequence of EEG-tokens. Thus, the Transformer receives a sequence of length $S+4$ during pretraining. The metadata dropout is applied with probability $p_{\mathrm{meta}}$ to reduce excessive dependence on the recording context. When a metadata field is dropped, its identifier is replaced with the corresponding unknown identifier, or its prefix embedding is set to zero.

The encoder applies multi-head self-attention, feed-forward layers with GELU activation, residual connections, and layer normalization. After encoding, the metadata-prefix outputs are discarded, and the EEG-token hidden states are used for masked code prediction and downstream representation.

\subsection{Masked Code Prediction Objective}

Although the Transformer encoder is shared across all frequency bands, each band has a separate prediction head because each VQ-VAE tokenizer has its own codebook and vocabulary size. Let $\mathbf{h}_i$ denote the final Transformer hidden state at EEG-token position $i$ after removing the metadata-prefix positions. For band $b$, the prediction head maps $\mathbf{h}_i$ to logits over the local band-specific vocabulary:
\begin{equation}
    \boldsymbol{\ell}^{(b)}_i
    =
    \mathbf{W}_{b}\mathbf{h}_i+\mathbf{a}_{b},
\end{equation}
where $\mathbf{W}_{b}\in\mathbb{R}^{K_b\times d}$ and $\mathbf{a}_{b}\in\mathbb{R}^{K_b}$.

For storage and input embedding, the VQ-code indices across all bands are represented in a unified global vocabulary, with band-specific offsets. Let $o_b$ denote the global offset assigned to band $b$. For a global target code index $c_i$ from band $b$, the corresponding local target index is
\begin{equation}
    \tilde{c}_i=c_i-o_b,
\end{equation}
where $\tilde{c}_i\in\{0,\dots,K_b-1\}$.

The Transformer encoder is pretrained to predict the original VQ code index at masked EEG-token positions. The masked code prediction loss for a sequence from band $b$ is
\begin{equation}
    \mathcal{L}_{\mathrm{MCP}}
    =
    \frac{
    \sum_{i=1}^{S}
    m_i\,
    \mathrm{CE}\!\left(
    \boldsymbol{\ell}^{(b)}_i,\tilde{c}_i
    \right)
    }{
    \sum_{i=1}^{S}m_i
    },
\end{equation}
where $m_i\in\{0,1\}$ indicates whether EEG-token position $i$ is masked, and $\mathrm{CE}(\cdot,\cdot)$ denotes the cross-entropy loss over the local vocabulary of band $b$.

\subsection{Downstream Representation}

For downstream classification, each trial $\mathbf{X}\in\mathbb{R}^{C\times N}$ is converted into five band-wise token streams using the trained VQ-VAE tokenizers. The trial is reflect padded, decomposed into the five frequency bands, center cropped to the original interval, segmented into temporal tokens, and encoded by the corresponding band-specific tokenizer. For each band, the resulting VQ code indices are offset by the band-specific global offset and flattened in the same time-major order used during pretraining. The corresponding quantized log-power tokens, channel identifiers, temporal-position indices, and metadata tokens are constructed using the same procedure as in pretraining.

The supervised classifier consists of the pretrained Transformer encoder followed by a trial-level classification head. For each trial, the five band-specific token sequences are encoded separately by the shared Transformer. Let $\mathbf{H}^{(b)}\in\mathbb{R}^{S\times d}$ denote the final EEG-token hidden states for band $b$ after removing the metadata-prefix outputs, where $S=CT$ and $d$ is the Transformer hidden dimension. The hidden states from all five bands are concatenated along the sequence dimension:
\begin{equation}
    \mathbf{H}_{\mathrm{trial}}
    =
    \mathrm{Concat}_{b\in\{\delta,\theta,\alpha,\beta,\gamma\}}
    \mathbf{H}^{(b)}
    \in\mathbb{R}^{5S\times d}.
\end{equation}
The trial-level representation is obtained by flattening $\mathbf{H}_{\mathrm{trial}}$, and a two-layer multilayer perceptron maps the resulting vector to class logits. The encoder and classification head are fine-tuned jointly using standard cross-entropy loss.

\section{Experiments}

\subsection{Pretraining Corpus}

We use the same compilation of 71 public EEG datasets introduced in SingLEM~\cite{ref_SingLEM} for self-supervised pretraining. The corpus contains more than 10,200 hours of multi-channel recordings, corresponding to more than 357,000 single-channel hours from more than 9,200 subjects. It covers diverse tasks, montages, references, sampling rates, file formats, recording durations, subject populations, and acquisition devices. This diversity supports the objective of learning general-purpose EEG representations rather than features tied to a single task, dataset, or acquisition system.

\subsection{Downstream Datasets}

We evaluate downstream transfer on the six EEG classification datasets used in the SingLEM benchmark~\cite{ref_SingLEM}. The benchmark contains three motor imagery datasets and three cognitive-task datasets. The motor imagery tasks include two classes: left versus right hand imagery and three classes: left hand, right hand, and foot imagery. The cognitive tasks include working memory, discrimination/selection responses, and word-generation paradigms. Table~\ref{tab:downstream_datasets} summarizes the downstream datasets.

\begin{table*}[t]
\centering
\caption{Summary of downstream evaluation datasets. Trials denotes the number of trials per subject.}
\label{tab:downstream_datasets}
\setlength{\tabcolsep}{4pt}
\renewcommand{\arraystretch}{1.08}
\begin{tabular}{l l c c c c c}
\hline
Dataset & Task type & Subjects & Classes & Trials & Channels & Trial duration \\
\hline
Dreyer~\cite{ref_dreyer} & Motor imagery & 21 & 2 & 160 & 27 & 5 s \\
WBCIC-2~\cite{ref_wbcic} & Motor imagery & 51 & 2 & 200 & 59 & 4 s \\
WBCIC-3~\cite{ref_wbcic} & Motor imagery & 11 & 3 & 300 & 59 & 4 s \\
N-back~\cite{ref_shin} & Cognitive & 26 & 2 & 108 & 28 & 10 s \\
DSR~\cite{ref_shin} & Cognitive & 26 & 2 & 72 & 28 & 10 s \\
WG~\cite{ref_shin} & Cognitive & 26 & 2 & 60 & 28 & 10 s \\
\hline
\end{tabular}
\end{table*}

\subsection{Baselines and Comparison Setting}

We compare BandVQ with pretrained EEG foundation models, including SingLEM~\cite{ref_SingLEM}, BENDR~\cite{ref_BENDR}, BIOT~\cite{ref_BIOT}, LaBraM~\cite{ref_LaBraM}, and CBraMod~\cite{ref_CBraMod}. The baseline scores are taken from the SingLEM benchmark tables. In that benchmark, the pretrained models were used as fixed feature extractors, and the extracted representations were classified using an SVM under a leave-one-subject-out protocol. Each baseline was adapted to its required montage by selecting the required channels or, when necessary, substituting missing channels with the physically closest available electrodes.

In contrast, the proposed model is fully fine-tuned with a classification head. Therefore, the comparison is not a strictly controlled architecture-only comparison under identical adaptation settings. Instead, it shows how the proposed fully fine-tuned band-wise model compares with published EEG foundation model results on the same downstream datasets and subject-independent splits. We account for this difference when interpreting the results.

\subsection{Evaluation Protocol}

We use leave-one-subject-out cross-validation to evaluate subject-independent transfer. For each fold, one subject is held out for testing, and the remaining subjects are used for training and validation. The non-test subjects are split at the subject level into an inner training set and a validation set. In each fold, 20\% of the non-test subjects are used for validation. The validation set is used for model selection and early stopping. After selecting the best epoch, the model is retrained on the full non-test set using the selected training duration and evaluated once on the held-out subject.

This protocol prevents trial leakage across subjects and evaluates generalization to unseen subjects. We report the mean performance across leave-one-subject-out folds.

\subsection{Training and Implementation Details}

All EEG recordings are resampled to 128 Hz and segmented into non-overlapping 1-s tokens, yielding $L=128$ samples per token. The signal scaling factor is $s=100$. For RMS normalization, $\epsilon=0.01$ and $\rho=0.01$. Absolute log-power values are computed with $\epsilon_p=10^{-8}$, clipped to $[-0.1,4.0]$, and uniformly quantized into 128 bins. The VQ latent dimension is $D=32$, the EMA decay for codebook updates is 0.99, and the codebook sizes for the delta, theta, alpha, beta, and gamma tokenizers are $(512,512,512,768,1024)$, with global offsets $(0,512,1024,1536,2304)$.

Each band-specific VQ-VAE tokenizer is trained independently with a batch size of 512, a learning rate of $3\times10^{-4}$, a weight decay of $10^{-5}$, and gradient clipping at 1.0. The tokenizer loss weights are $(\lambda_{\mathrm{MAE}},\lambda_{\mathrm{MSE}},\lambda_{\mathrm{STFT}},\lambda_{\mathrm{commit}})=(1.25,0.5,0.25,1.0)$, and the STFT loss uses $n_{\mathrm{fft}}\in\{256,512,1024\}$ with a hop length of $0.25n_{\mathrm{fft}}$.

The Transformer has 12 layers, a model dimension of $d=256$, and 8 attention heads. It is pretrained with AdamW using a batch size of 16, a peak learning rate of $3\times10^{-4}$, a minimum learning rate of $3\times10^{-5}$, weight decay of $10^{-2}$, gradient clipping at 1.0, cosine scheduling, and 10,000 warmup updates. Region masking uses $p_{\mathrm{mask}}=0.5$, and metadata dropout uses $p_{\mathrm{meta}}=0.1$.

For downstream classification, the encoder and classification head are fine-tuned jointly with a batch size of 16. The encoder and classifier learning rates are $3\times10^{-5}$ and $10^{-3}$, respectively. The classifier uses a hidden size of 256 and a dropout of 0.2. Training runs for at most 30 epochs, with a minimum of 6 epochs and an early-stopping patience of 8. The implementation uses Python 3.12.4, PyTorch 2.4.1, CUDA 12.1, and four A100 80 GB GPUs.

\subsection{Evaluation Metrics}

We report accuracy and the macro-averaged $F_1$ score. Accuracy measures the overall proportion of correctly classified trials, whereas the macro-$F_1$ averages the class-wise $F_1$ scores and therefore gives equal weight to each class.

\subsection{Ablation Settings}

We conduct ablation studies to assess the contribution of the key conditioning components in the proposed framework. The full model uses both metadata prefix tokens and quantized band-power tokens. The first ablation removes the metadata prefix tokens while retaining the band-power tokens. The second ablation removes both metadata and band-power tokens. These settings isolate the effect of contextual metadata and test whether the amplitude information restored by power tokens improves downstream transfer after RMS-normalized VQ-VAE tokenization.

\section{Results}

\subsection{Comparison on Motor Imagery Tasks}

Table~\ref{tab:MI_results} reports the classification results on the three motor imagery datasets. BandVQ achieves the second-best performance on all three datasets. It obtains 72.92\% accuracy and 72.66\% macro-$F_1$ on Dreyer, 66.29\% accuracy and 65.12\% macro-$F_1$ on WBCIC-3, and 79.19\% accuracy and 79.02\% macro-$F_1$ on WBCIC-2.

SingLEM achieves the best reported results on the motor imagery tasks. BandVQ remains close to SingLEM, particularly on WBCIC-2, where the accuracy gap is 0.36 percentage points. Under the reported benchmark, fine-tuned BandVQ also outperforms the multi-channel baselines BENDR, BIOT, LaBraM, and CBraMod across all three motor imagery datasets. These results indicate that the band-wise vector-quantized representation provides competitive transfer performance for subject-independent motor imagery classification, although it does not surpass the strongest single-channel baseline.

\begin{table*}[t]
\centering
\caption{Classification results on three motor imagery datasets. Values denote mean performance. Bold indicates the best result, and underline indicates the second-best result for each metric and dataset.}
\label{tab:MI_results}
\setlength{\tabcolsep}{9pt}
\begin{tabular}{@{}l@{\hspace{5pt}}cc@{\hspace{8pt}}cc@{\hspace{8pt}}cc@{}}
\toprule
\multirow{2}{*}{\textbf{Model}} 
& \multicolumn{2}{c}{\textbf{Dreyer}} 
& \multicolumn{2}{c}{\textbf{WBCIC-3}} 
& \multicolumn{2}{c}{\textbf{WBCIC-2}} \\
\cmidrule(lr){2-3} \cmidrule(lr){4-5} \cmidrule(l){6-7}
& Acc. (\%) & F1 (\%) & Acc. (\%) & F1 (\%) & Acc. (\%) & F1 (\%) \\ 
\midrule
BENDR \cite{ref_BENDR}  & 50.51 & 50.24 & 35.28 & 34.84 & 51.32 & 51.24 \\
BIOT \cite{ref_BIOT}   & 56.85 & 52.64 & 37.04 & 33.78 & 51.20 & 48.25 \\
LaBraM \cite{ref_LaBraM} & 49.91 & 34.02 & 40.04 & 37.17 & 56.88 & 56.15 \\
CBraMod \cite{ref_CBraMod} & 71.10 & 70.62 & 59.93 & 58.96 & 77.56 & 77.26 \\
SingLEM \cite{ref_SingLEM}
& \textbf{75.27} & \textbf{75.16} 
& \textbf{68.26} & \textbf{68.17} 
& \textbf{79.55} & \textbf{79.45} \\
\midrule
\textbf{BandVQ}
& \underline{72.92} & \underline{72.66} 
& \underline{66.29} & \underline{65.12} 
& \underline{79.19} & \underline{79.02} \\
\bottomrule
\end{tabular}
\end{table*}

\subsection{Comparison on Cognitive Tasks}

Table~\ref{tab:Cognitive_results} reports the results on the three cognitive task datasets. BandVQ achieves the best performance on all three tasks. Compared with SingLEM, the strongest baseline in the table, the proposed model improves accuracy by 2.24 percentage points on N-back, 1.28 percentage points on DSR, and 3.34 percentage points on WG. The corresponding macro-$F_1$ improvements are 2.57, 1.06, and 2.75 percentage points, respectively.

The improvement is consistent across the three cognitive datasets, suggesting that the proposed representation is effective for tasks involving distributed and frequency-dependent neural dynamics. The gains over CBraMod and the larger gains over BENDR, BIOT, and LaBraM also suggest that band-wise discrete tokenization, metadata conditioning, and explicit power representation are useful for transferring to cognitive EEG classification tasks.

\begin{table*}[t]
\centering
\caption{Classification results on three cognitive task datasets. Values denote mean performance. Bold indicates the best result, and underline indicates the second-best result for each metric and dataset.}
\label{tab:Cognitive_results}
\setlength{\tabcolsep}{9pt}
\begin{tabular}{@{}l@{\hspace{5pt}}cc@{\hspace{8pt}}cc@{\hspace{8pt}}cc@{}}
\toprule
\multirow{2}{*}{\textbf{Model}} 
& \multicolumn{2}{c}{\textbf{N-back}} 
& \multicolumn{2}{c}{\textbf{DSR}} 
& \multicolumn{2}{c}{\textbf{WG}} \\
\cmidrule(lr){2-3} \cmidrule(lr){4-5} \cmidrule(l){6-7}
& Acc. (\%) & F1 (\%) & Acc. (\%) & F1 (\%) & Acc. (\%) & F1 (\%) \\
\midrule
BENDR \cite{ref_BENDR}
& 52.96 & 52.75 
& 54.59 & 54.43 
& 52.24 & 51.94 \\

BIOT \cite{ref_BIOT}
& 58.90 & 54.66 
& 60.84 & 57.91 
& 56.67 & 52.54 \\

LaBraM \cite{ref_LaBraM}
& 62.25 & 59.84 
& 66.83 & 65.60 
& 60.64 & 56.44 \\

CBraMod \cite{ref_CBraMod}
& {78.13} & {76.81} 
& {79.59} & {78.81} 
& {69.36} & {67.93} \\

SingLEM \cite{ref_SingLEM}
& \underline{82.34} & \underline{81.65} 
& \underline{84.72} & \underline{84.50} 
& \underline{69.87} & \underline{69.58} \\

\midrule
\textbf{BandVQ}
& \textbf{84.58} & \textbf{84.22} 
& \textbf{86.00} & \textbf{85.56} 
& \textbf{73.21} & \textbf{72.33} \\
\bottomrule
\end{tabular}
\end{table*}

\subsection{Ablation Results}

Table~\ref{tab:ablation_results} reports the ablation results across all six downstream datasets. Removing metadata decreases accuracy on five of the six datasets. The accuracy drops are 3.01 percentage points on Dreyer, 1.78 on WBCIC-3, 0.60 on WBCIC-2, 0.61 on N-back, and 1.92 on DSR. WG is the only exception, where removing metadata improves accuracy from 73.21\% to 74.29\%.

Removing both metadata and band-power tokens causes larger and more consistent degradation. Compared with the full model, accuracy decreases by 12.71 percentage points on Dreyer, 6.82 points on WBCIC-3, 2.71 points on WBCIC-2, 10.08 points on N-back, 8.22 points on DSR, and 7.57 points on WG. These results show that the quantized band-power tokens provide a substantial contribution to downstream transfer. The large gap between the metadata-only ablation and the metadata plus power ablation indicates that amplitude information is particularly important after RMS-normalized tokenization.

\begin{table*}[t]
\centering
\caption{Ablation study across six datasets. Values denote mean accuracy. Bold indicates the best result, and underline indicates the second-best result for each dataset.}
\label{tab:ablation_results}
\setlength{\tabcolsep}{6pt}
\begin{tabular}{@{}l@{\hspace{6pt}}cccccc@{}}
\toprule
\textbf{Method} 
& \textbf{Dreyer} 
& \textbf{WBCIC-3} 
& \textbf{WBCIC-2} 
& \textbf{N-back} 
& \textbf{DSR} 
& \textbf{WG} \\
\midrule
Full 
& \textbf{72.92} 
& \textbf{66.29} 
& \textbf{79.19} 
& \textbf{84.58} 
& \textbf{86.00} 
& \underline{73.21} \\

w/o metadata 
& \underline{69.91} 
& \underline{64.51} 
& \underline{78.59} 
& \underline{83.97} 
& \underline{84.08} 
& \textbf{74.29} \\

w/o meta+power 
& 60.21 
& 59.47 
& 76.48 
& 74.50 
& 77.78 
& 65.64 \\
\bottomrule
\end{tabular}
\end{table*}

\section{Discussion}
The results show that the proposed band-wise vector-quantized model, BandVQ, provides strong downstream performance across heterogeneous subject-independent EEG classification tasks. The model is particularly effective on cognitive tasks, suggesting that frequency-specific waveform structure and explicit band-power information may benefit tasks whose discriminative information is distributed across multiple frequency bands. On motor imagery tasks, the model remains competitive but does not surpass SingLEM, possibly because sensorimotor rhythms and subject-specific spatial patterns are already effectively captured by strong single-channel representations.

The ablation results support the importance of the proposed conditioning components. Removing metadata decreases accuracy on five of six datasets, whereas removing both metadata and band-power tokens results in larger and more consistent degradation, supporting the separation of normalized waveform morphology from absolute amplitude information. The WG result, showing that metadata removal improves accuracy, suggests that metadata effects may depend on the task or annotation quality.

Several limitations remain. First, the comparison with existing foundation models is not fully controlled, as the proposed model is fine-tuned, whereas the baseline scores are from a fixed-feature benchmark. Second, additional ablations are needed to isolate the effects of region-based masking, band-specific codebook sizes, metadata dropout, and band-fusion strategies. Third, evaluations of clinical EEG, sleep staging, affective computing, seizure-related tasks, and cross-dataset evaluation would provide a broader assessment of general-purpose EEG representation learning.



\section{Conclusion}
This paper presented \textbf{BandVQ}, a band-wise vector-quantized EEG foundation model that pairs five frequency-specific VQ-VAE tokenizers with a shared Transformer encoder pretrained by masked code prediction. Metadata prefix tokens, quantized log-power tokens, and region-based masking address spectral imbalance, acquisition heterogeneity, and spatial redundancy in EEG. On six subject-independent downstream tasks, the model achieved the best reported accuracy on three cognitive datasets and competitive results on three motor imagery datasets. These findings suggest that separating frequency-specific tokenization from shared contextual representation learning is a promising direction for scalable EEG foundation models.
%
%
%
%

\bibliographystyle{splncs04}
\bibliography{references_nodoi}

\end{document}